\documentclass{article}
\usepackage{spconf,amsmath,graphicx}
\usepackage{xcolor}

\title{HSTR-NET: HIGH SPATIO-TEMPORAL RESOLUTION VIDEO GENERATION \\ FOR WIDE AREA SURVEILLANCE}
\twoauthors
  {H. Umut Suluhan \thanks{This work is supported in part by TUBITAK Grant Project No: 118E891}}
	{Ozyegin University\\
	Computer Science Department\\
	34794 Istanbul, Turkey}
  {Hasan F. Ates, Bahadir K. Gunturk} 
	{Istanbul Medipol University\\
	School of Engineering and Natural Sciences\\
	 34810 Istanbul, Turkey}

\begin{document}
%\ninept
%
\maketitle
\begin{abstract}
Wide area surveillance has many applications and tracking of objects under observation is an important task, which often needs high spatio-temporal resolution (HSTR) video for better precision.
This paper presents the usage of multiple video feeds for the generation of HSTR video as an extension of reference based super resolution (RefSR). One feed captures video at high spatial resolution with low frame rate (HSLF) while the other captures low spatial resolution and high frame rate (LSHF) video simultaneously for the same scene. The main purpose is to create an HSTR video from the fusion of HSLF and LSHF videos. 
In this paper we propose an end-to-end trainable deep network 
%motivated by RIFE\cite{huang2020rife} architecture 
that performs optical flow estimation and frame reconstruction by combining inputs from both video feeds. The proposed architecture provides significant improvement over existing video frame interpolation and RefSR techniques in terms of objective PSNR and SSIM metrics.
\end{abstract}
\section{Introduction}
\label{sec:intro}

Wide area surveillance (WAS) requires wide angle high resolution cameras that continuously survey vast regions, preferably at a high frame rate. Professional WAS camera systems are both bulky and expensive. Hence, alternate ways of acquisition of high spatio-temporal resolution (HSTR) video has been an active area of research.  Computational imaging techniques can be applied to produce HSTR videos using multiple lower resolution cameras \cite{Dual_Camera}. Resolution enhancement has been extensively studied in the context of both spatial super resolution (SR), when high resolution (HR) frames are synthesized from corresponding low resolution (LR) frames, and also temporal frame interpolation (TFI) for improving the temporal resolution of low frame rate videos.
However, estimation of missing information spatially or temporally is a challenging task since (i) TFI creates motion blur and artifacts  due to assumption of linear motion among the neighboring frames, and (ii) spatial interpolation creates over-smoothed frames missing necessary high frequency components \cite{xue2019video}.

Applying single image super-resolution (SISR) directly to each LR video frame produces HR video but lacks temporal coherency. To compensate for that, temporal information can be incorporated into video SR process, as sequence of neighboring frames are motion compensated and used to super-resolve the given frame. The quality of video SR can be significantly improved if we provide some HR reference image during the SR process to construct the high frequency details. Recent stream of reference-based super resolution
(RefSR) networks have been introduced as an extension of multiple cameras combined in hybrid systems \cite{Dual_Camera, zheng2018crossnet, zhang2019image}. Using multiple video feeds, SR is performed based on some well-registered reference imagery \cite{ xie2020feature, zheng2017learning, dong2021rrsgan, zhao2018cross, gigapixel}. In this paper, we apply this reference-based approach in the domain of temporal frame interpolation with the assistance of corresponding LR frames that are captured by a second video feed. 

\begin{figure*}
\begin{center}
\includegraphics[width=\textwidth]{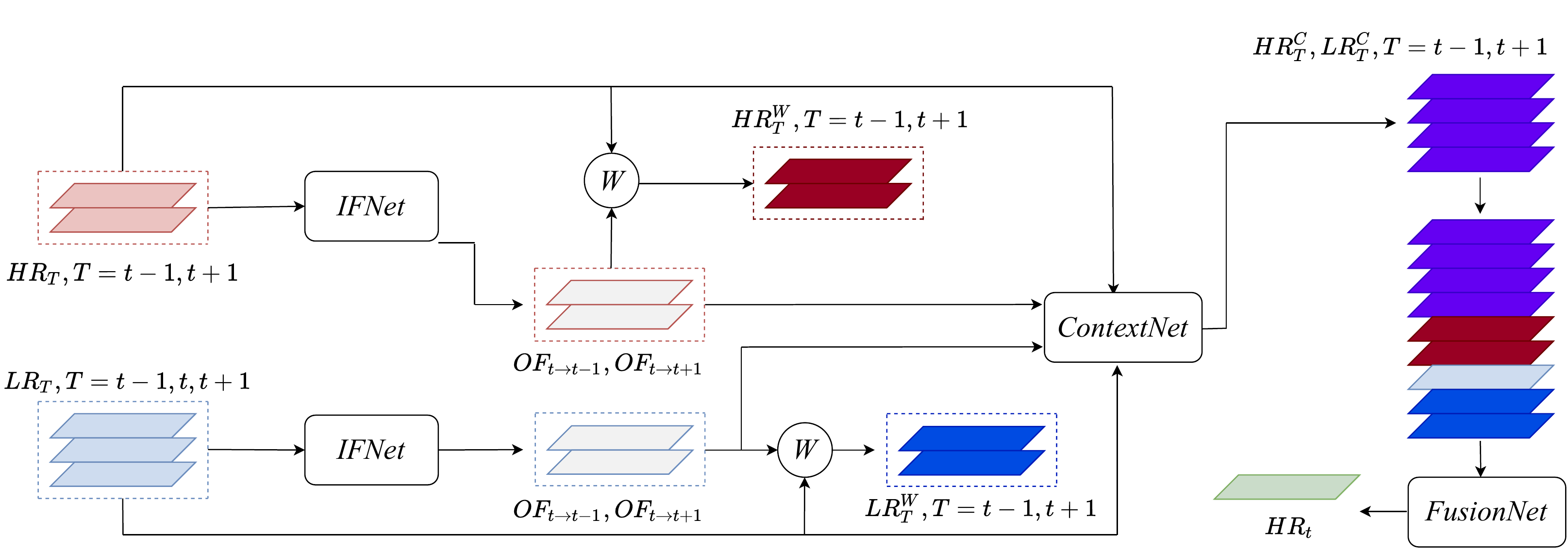}
\end{center}
\vspace{-0.5cm}
   \caption{The block diagram of proposed HSTR-Net.}
\label{fig:block_diagram}
\vspace{-0.3cm}
\end{figure*}

Frame rate conversion is referred to as synthetically adding missing frames to increase the frame rate by typically assuming linear motion among the neighboring frames. Recently deep network architectures are used to estimate optical flow \cite{huang2020rife,xue2019video} and to construct the missing frame, such as the popular channel attention based models
\cite{choi2020channel,bao2019depth,shen2020blurry}. But motion between frames is not always linear, which adds motion induced artifacts, and estimating better motion vectors is still a challenging task. The proposed scenario deals with the preliminary limitation in TFI due to linear motion assumption. In our proposed methodology, frame HR$_t$ can be synthesized between frames  HR$_{t-1}$ and HR$_{t+1}$, even when there is large and irregular motion, with assistance from corresponding LR$_{\tau}$, $(t-1\leq \tau \leq t+1)$, frames that are used as reference. Hence, overall we want to synthesize HSTR$(t, h, w, f )$ video from HSLF$(t, h, w, f/p )$ and LSHF$(t, h/r, w/r, f )$ videos, where $h, w, r, f, p $ are the height, width, scale factor, frame rate (fps) and frame rate factor, respectively. 

In this paper we provide an end-to-end trainable deep CNN-based network (named as HSTR-Net) to synthesize HSTR video using HSLF and LSHF video feeds. Proposed method uses two HR and three LR frames as input, extracts the necessary high-frequency components from HR inputs and uses LR inputs as reference for better motion compensation and artifact-free reconstruction. The block diagram for proposed architecture is provided in Figure \ref{fig:block_diagram} and detailed explanation is given in the Section \ref{sec:proposed method}. Simulation results in Section \ref{sec:experiments and results} show that HSTR-Net model is superior to state-of-the-art RefSR and TFI methods in terms of PNSR and SSIM metrics and also has low computational complexity. The main contributions of our work can be summarized as:
\begin{itemize}
\item We provide an end-to-end trainable network for synthesizing an HR frame from neighboring HR frames under the assistance of corresponding LR frames. 
\item We introduce modules for low complexity optical flow estimation and trainable warping with deformable convolutions.
\item HSTR-Net is robust against complex and large motion, as typically seen in aerial surveillance videos, and is suitable for real-time processing in embedded GPUs, e.g. on a power-limited drone.
\end{itemize}

\begin{figure*}
\begin{center}
\includegraphics[width=\textwidth]{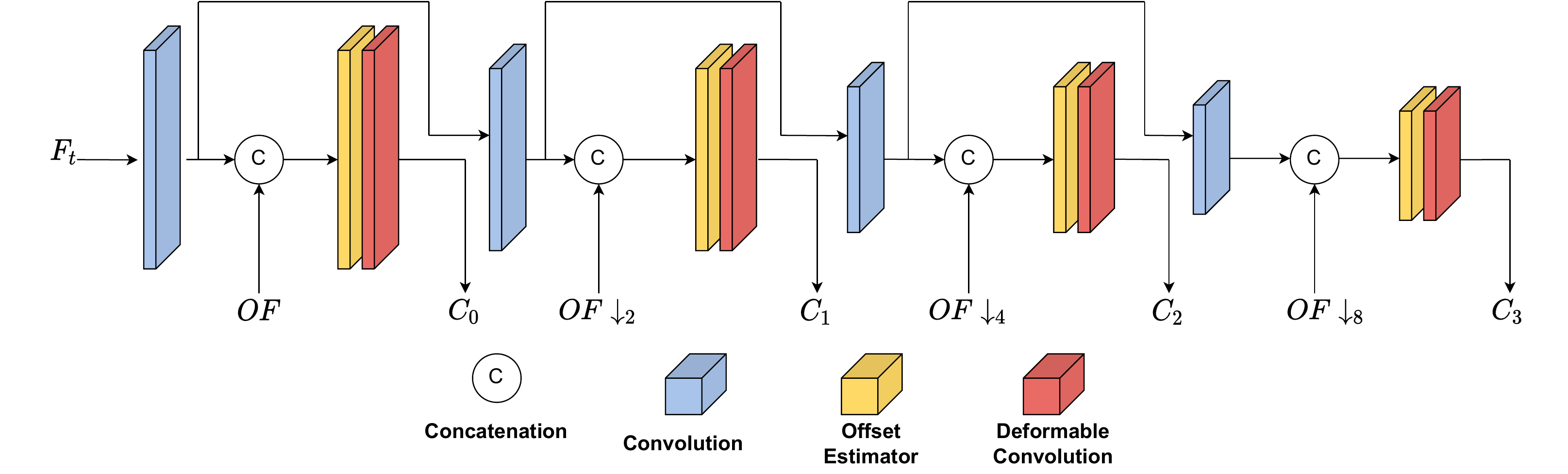}
\end{center}
   \vspace{-0.7cm}
   \caption{ContextNet architecture. $F_t$ and $OF$ denote input frame and optical flow respectively. In each level input feature map and $OF$ is downscaled by a factor of 2. $C_0$, $C_1$, $C_2$ and $C_3$ denote differently sized outputs of ContextNet. %\textcolor{red}{HFA: OF'in downscale olduğu bu resimden anlaşılmıyor. HUS:Diyagramı büyüterek yapabilirim hocam.}
   }
\label{fig:contextnet_diagram}
\vspace{-0.3cm}
\end{figure*}

\vspace{-0.4cm}
\section{PROPOSED ARCHITECTURE}
\label{sec:proposed method}
\vspace{-0.2cm}
In this section we introduce the novel HSTR-Net architecture, a reference based interpolation method to synthesize the high spatio-temporal video. We propose a fully convolutional end-to-end trainable network and baseline approach for proposed methodology. The task under consideration is to mix the two video feeds, one being LSHF (e.g. $112 \times 64$ spatial resolution and $30 fps$) and other HSLF (e.g. $448 \times 256$ having $15fps$) to synthesize one HSHF video (with $448 \times 256$ and $30 fps$). To synthesize frame HR$_{t}$ between  frames  HR$_{t-1}$ and HR$_{t+1}$, we have used total of 5 frames from the two video sequences as LR$_\tau$, $(t-1\leq \tau\leq t+1)$ and HR$_\tau$, $(\tau=t-1,t+1)$. The proposed architecture in Figure \ref{fig:block_diagram} is motivated by RIFE\cite{huang2020rife}, which is a deep frame interpolation  network.
\vspace{-0.4cm}
\subsection{IFNet}
%\textcolor{red}{HFA: IFNet ile ilgili biraz teknik detay gerekli}\\
%\textcolor{red}{HUS: Eklediğim ve değiştirdiğim kısımları kırmızı ile belirttim hocam.}
\label{head: ifnet}
\vspace{-0.1cm}
IFNet is the  input block of HSTR-Net and is used to extract the optical flow (OF) between input frames. We have used the flow estimation module  originally proposed in \cite{huang2020rife} for HSTR-Net. IFNet consists of three IFBlocks that serve as pipeline for the optical flow estimation. Each IFBlock is composed of sequential convolutional layers with 64 channels. Previously estimated flow is upsampled by a factor of 2 before each IFBlock to extract flow information at different resolutions and to capture large motions as well. We have used IFNet to predict the optical flow from neighboring frames to the center frame. At the output of IFNet, we obtain following motion vector fields for both HR and LR frames:
$$OF_{t \rightarrow t-1}, OF_{t \rightarrow t + 1}$$
\vspace{-1.0cm}
\subsection{Warping}
\label{head: warping}
\vspace{-0.1cm}
Warping is the image registration module used to register one frame on to the other with the help of optical flow. We have used back-warping module as explained in \cite{huang2020rife}. This module initially splits flow into vertical and horizontal components, performs sub-pixel motion compensation on the input frame to align the input frame with the targeted output.

%\begin{figure}[htb]
%\begin{minipage}[b]{1.0\linewidth}
%  \centering
%  \centerline{\includegraphics[width=8.5cm]{ICIP/images/contextnet_deformconv.pdf}}
%\end{minipage}
%\caption{ContextNet Architecture}
%\label{fig:contextnet_diagram}
%\end{figure}
\vspace{-0.4cm}

\subsection{ContextNet}
\label{head: contextnet}
%\textcolor{red}{HFA: offset estimator ve deformable conv. ile ilgili biraz teknik detay gerekli}\\
%\textcolor{red}{HUS: Eklediğim ve değiştirdiğim kısımları kırmızı ile belirttim hocam.}
\vspace{-0.1cm}

ContextNet is a feature extraction model inspired from \cite{huang2020rife}. ContextNet gradually lowers the resolution of input frame and flows, and then uses warping module to back-warp flow features on to the frame to be synthesized. By doing that, ContextNet generates warped feature channels at different resolutions that are later used to  reconstruct accurately the HR$_{t}$ frame. However, warping is not trainable, hence cannot be adjusted based on content. We replace warping with offset estimator and deformable convolution layers \cite{shim2020robust}, in order to be able to fine-tune registration process of flows on to frames (see Figure \ref{fig:contextnet_diagram}). Deformable convolution gives better results in terms of PSNR/SSIM metrics, but slows down HSTR-Net. 

Figure \ref{fig:contextnet_diagram} illustrates the modified ContextNet architecture. ContextNet produces feature maps, $C_i$, at 4 different scales by downscaling the input frame $F_t$ through convolutional layers with a stride of 2.  Optical flow is also downsampled to match the feature map size. The convolution layer output is concatenated with the optical flow and given as input to the offset estimator. Input features and output of the offset estimator go through deformable block for warped feature map generation. Offset estimator dynamically calculates filter offset which is used for spatially varying (i.e. deformable) convolution. In \cite{shim2020robust} HR reference and LR frame features are used as input for offset estimator, whereas we give reference frame features and optical flow to the offset estimator. Hence in  ContextNet, offset estimator is used to improve the accuracy of optical flow vectors and provide more precise registration of feature maps. Offset estimator module is composed of 3 downsample, 3 attention and upsample blocks, and one convolutional layer to scale the channels. Deformable convolution takes advantage of trainable offset to compute registered feature maps. In HSTR-Net, deformable block has one convolutional and one deformable convolution layer to process the input features and the offset. The number of channels  in feature maps $C_0, C_1, C_2, C_3$ are 16, 32, 64 and 128, respectively.

%\begin{table}
%\begin{center}
%\begin{tabular}{||c c c c||} 
% \hline
% Experiment & PSNR & SSIM & TIME(ms) \\
% \hline
% \hline
%HSTR-Net & 40.14 & 0.988 & 15.8  \\ 
%\hline
%HSTR-Net* & 40.62 & 0.989 & 60.8  \\ 
%\hline
%\end{tabular}
%\caption{Comparison of HSTRNet models with and without deformable %convolution on Vimeo90K\cite{xue2019video} Dataset(* denotes model with %deformable convolution\cite{shim2020robust})}
%\end{center}
%\end{table} 
%
\vspace{-0.4cm}

\subsection{FusionNet}
%\textcolor{red}{HFA: FusionNet ile ilgili biraz teknik detay gerekli}\\
%\textcolor{red}{HUS: Eklediğim ve değiştirdiğim kısımları kırmızı ile belirttim hocam.}
\label{head:fusionnet}

\vspace{-0.2cm}

FusionNet is a reconstruction block that is inspired from UNet\cite{ronneberger2015u} auto-encoder structure to infer the HR$_{t}$ frame. FusionNet consists of an encoder-decoder network and residual connections to extract features at different resolutions which are later fused during frame reconstruction. In HSTR-Net, we give the following inputs to the FusionNet:
\begin{itemize}
\item Warped HR frames: ${HR}_{T}^{W}, T = t - 1, t + 1$
\vspace{-0.2cm}
\item Warped LR frames:${LR}_{T}^{W}, T = t - 1, t + 1$
\vspace{-0.2cm}
\item LR reference frame: ${LR}_{t}$
\vspace{-0.2cm}
\item ContextNet outputs: ${HR}_{T}^{C}, {LR}_{T}^{C}, T = t - 1, t + 1$
\end{itemize}

FusionNet consists of four downsampling at the encoder and four upsampling convolutional layers at the decoder. Additionally, one final convolutional layer is used to upsample the output and match the initial frame size. ContextNet outputs $C_i$ are concatenated with encoder layer outputs at different scales and given as input to the next layer. The number of output channels in downsampling and upsampling layers are (32, 96, 224, 480) and (128, 64, 32, 16), respectively. Last convolutional layer has 16 input and 4 output channels. 3 output channels correspond to the residual high-frequency components of the synthesized HR frame, and the other output is a pixel-level mask used for computing weighted average of the warped HR frames. Finally the reconstructed frame is computed as follows:
\vspace{-0.1cm}
\begin{equation}
 \hat{\mbox{HR}}_t = m\mbox{HR}^W_{t-1} + (1-m)\mbox{HR}^W_{t+1} + R
 \vspace{-0.1cm}
\end{equation}
where $R$ is the residual output and $m$ is the generated mask.
\vspace{-0.3cm}
\section{EXPERIMENTS AND RESULTS}
\label{sec:experiments and results}
\vspace{-0.2cm}
\subsection{Datasets}
\label{head: datasets}
\vspace{-0.1cm}
The proposed model is trained on the Vimeo90K\cite{xue2019video} dataset and tested on both Vimeo90K and Visdrone\cite{zhu2021viz} datasets. Vimeo90K dataset consists of 51313 training and 3782 test triplets that are recorded by commercial cameras in resolution of $448 \times 256$. Visdrone test dataset is generated from Visdrone aerial videos by resizing the frames to the resolution of $672 \times 380$ for a total of 2804 test samples. Visdrone provides a rich test set with aerial drone videos having both camera and object motion. Both datasets only include HR frames; to simulate two-camera system, LR frames are synthesized by first 4$\times$ downsampling the HR frames and then 4$\times$ upsampling the downsampled frames using bicubic interpolation. For evaluation, Peak-Signal-to-Noise-Ratio (PSNR) and Structural Similarity Index Measure (SSIM) are computed between reconstructed and ground truth HR frames. 

\vspace{-0.3cm}

\subsection{Training}
\label{head: training}

\vspace{-0.1cm}
HSTR-Net is trained on Vimeo90K training set for 200 epochs, with a learning rate of $10^{-4}$ for 100 epochs, and a learning rate to $10^{-5}$ for the remaining 100 epochs. Data augmentation, including random cropping, rotation and flipping, is used during training. Random cropping increases training set size and also speeds up the training process since frames are cropped to the size of 128x128 pixels. A rotation angle between -10 to 10 degrees is selected for each training sample. Horizontal flipping is applied with \%20 probability.

%\begin{minipage}[b]{0.85\linewidth}
%  \centerline{\includegraphics[width=7.5cm]{ICIP/images/vimeo_table.png}}
%  \centerline{\textbf{Table 2.} Comparative Results on Vimeo90K\cite{xue2019video} Dataset.}
%  \cite{Dual_Camera,  zheng2018crossnet} results are as reported in the respective papers.We %evaluated rest of the results on our benchmark. (The discripancy of SSIM results for and rest %of the papers is caused due to different SSIM evaluation code in different papers.)\medskip
%\end{minipage}

 For training the model and supervising the reconstruction,  $L_1$ loss function is preferred \cite{zhao2016loss} instead of typically used Mean Squared Loss (MSE or $L_2$ loss).  The use of $L_1$ loss slightly reduces the  PSNR but visual results are qualitatively better and sharper images are generated \cite{lim2017enhanced}.   
%\begin{equation}
%   L_1  = {\lVert I_{GT} -I_{SR}  \rVert}_1
%\end{equation}

\begin{table}
%\caption{Comparative Results on Vimeo90K\cite{xue2019video} Dataset, where $f$ stands for $finetuned$ models on the Vimeo90K dataset. \cite{Dual_Camera,  zheng2018crossnet} results are as reported in the respective papers.We evaluated rest of the results on our benchmark. *Times reported for AWNet and CrossNet are predicted from the information given in the \cite{Dual_Camera} }
\caption{Comparative Results on Vimeo90K Dataset. AWNet and CrossNet results are as reported in the respective papers. Execution times for AWNet and CrossNet are adopted from \cite{Dual_Camera}. Other results are generated on the benchmark test set.}
\vspace{-0.1cm}
\begin{center}
\begin{tabular}{l c c c} 
 \hline
 Method & PSNR & SSIM & TIME(ms) \\
 \hline
 \hline
 AWNet\cite{Dual_Camera} & 39.88 & 0.986 & $\sim$100 \\
 \hline
 CrossNet\cite{zheng2018crossnet} & 39.17 & 0.985 & $\sim$150\\
 \hline
RIFE\cite{huang2020rife} & 35.61 & 0.978 & 10.6  \\ 
\hline
RIFE(2T2R)\cite{huang2020rife} & 36.13 & 0.980 & 9.7  \\ 
\hline
ToFlow-interp\cite{xue2019video} & 37.41 & 0.976 & 124.7  \\ 
\hline
%MASA-SR\cite{lu2021masasr} & 33.38 & 0.952 & 46.7  \\ 
%\hline
MASA-SR \cite{lu2021masasr} & 35.08 & 0.970 & 39.3  \\ 
\hline
SRNTT\cite{zhang2019image} & 33.90 & 0.961 & 1480.3  \\ 
\hline
HSTR-Net w/o dc & 40.14 & 0.988 & 15.8  \\ 
\hline
\textbf{HSTR-Net w/ dc} & \textbf{40.62} & \textbf{0.989} & \textbf{60.8}  \\ 
\hline
\end{tabular}
\vspace{-0.6cm}
\label{table1}
\end{center}
\end{table}

\vspace{-0.3cm}

\subsection{Discussion of Results}
\label{head: results}

\vspace{-0.1cm}
We compare HSTR-Net with previous models on video frame interpolation and reference-based SR \cite{ Dual_Camera, xue2019video, zheng2018crossnet, zhang2019image,huang2020rife, lu2021masasr}. For comparison, we also implemented ToFlow-interp, which is a RefSR version of ToFlow\cite{xue2019video} that uses the same set of HR and LR reference frames to reconsruct the missing HR frame. All models except for SRNTT\cite{zhang2019image} are either trained or fine-tuned on Vimeo90K dataset for fair evaluation. 

%Unfortunately, only using HR frames are not enough to reconstruct %high-quality frames as seen in the results even though the model is %finetuned on the Vimeo90K\cite{xue2019video} dataset.

As seen from Tables \ref{table1} and \ref{table2}, HSTR-Net achieves state-of-the-art results on both Vimeo90K and Visdrone datasets. HSTR-Net successfully makes use of additional information extracted from the LR frames. Other methods fall short in the domain of large and complex motion estimation when only HR frames (i.e. RIFE) or one HR and one LR frame (i.e MASA-SR) are used for reconstruction. Using only HR frames is not sufficient to capture large and complex motions. Visdrone dataset includes drone captured frames, hence includes large motions and instant changes of motion as well. LR frames contain additional information to aid in capturing and estimating large and complex motions between frames. Despite being trained on Vimeo90K, HSTR-Net generates state-of-the-art results on Visdrone dataset as well, proving that HSTR-Net is a well-generalized network that can be applied in different scenarios.

%Most RefSR \cite{zhang2019image, lu2021masasr} methods struggle to achieve better results in the Visdrone dataset due to using only HR frames for super-resolution process. HSTR-Net achieves the best result among other models because of usage of LR frames as well for super-resolution and frame interpolation. 

%WE CAN REMOVE BELOW SHORTER VERSION IS COMMENTED IN CONCLUSION
%We demonstrated that incorporating low-resolution frames into network has shown great progress on reconstruction of the frame. We think that using two video feeds that are captured by different cameras(one HSLF, one LSHF) can generate high-quality videos and eliminate cost of using expensive cameras. 

Tables \ref{table1} and \ref{table2} also provide HSTR-Net results with and without deformable convolution. The use of trainable offset estimator and deformable convolution in exchange for fixed warping module improves PSNR/SSIM results in both Vimeo90K and Visdrone datasets. However the inference time of the model is also significantly increased with the use of deformable convolution. Drones require low memory, fast and less computationally demanding networks for real-time processing due to limitations of computational power of embedded GPUs. HSTR-Net based on warping instead of deformable convolution provides a light-weight alternative for embedded real-time processing and achieves competitive results on reconstructed image quality.

%Warping based HSTR-Net requires lower computional power and is faster than deformable convolution based HSTR-Net because of its nonconvolutional architecture. Hence, warping based HSTR-Net is more convenient for real-time processing with embedded GPU drones.

%Deformable convolution raises both PSNR and SSIM results because of its trainability. Warping is not trainable to be fine-tuned further. However, deformable convolution has layers that are trainable to generate highly-accurate warped frames. 

\begin{table}
\caption{ Comparative Results on Visdrone  Dataset} %where $f$ stands for $finetuned$ models on the %Vimeo90K\cite{xue2019video} dataset. All results are obtained on our %benchmark.}
\vspace{0.1cm}
\begin{center}
\begin{tabular}{l c c c} 
 \hline
 Method & PSNR & SSIM & TIME(ms) \\
 \hline
 \hline
RIFE\cite{huang2020rife} & 33.53 & 0.962 & 8.7  \\ 
\hline
RIFE(2T2R)\cite{huang2020rife} & 33.63 & 0.963 & 8.1  \\ 
\hline
ToFlow-interp\cite{xue2019video} & 33.69 & 0.965 & 150.3  \\ 
\hline
MASA-SR\cite{lu2021masasr} & 24.53 & 0.830 & 110  \\ 
\hline
SRNTT\cite{zhang2019image} & 25.63 & 0.874 & 5661  \\ 
\hline
HSTR-Net w/o dc & 34.23 & 0.968 & 16.5  \\ 
\hline
\textbf{HSTR-Net w/ dc} &  \textbf{35.06} & \textbf{0.973} & \textbf{77.7}  \\ 
\hline
\end{tabular}
\label{table2}
\end{center}
\vspace{-0.6cm}
\end{table}

\begin{figure}[htb]
\vspace{-0.1cm}
\begin{minipage}[b]{1.0\linewidth}
  \centering
  \centerline{\includegraphics[width=8.5cm]{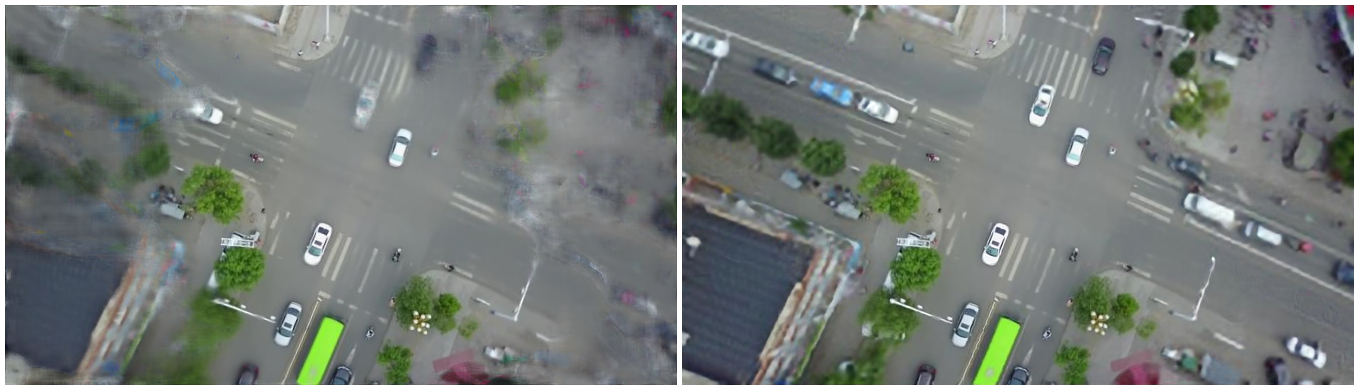}}
\end{minipage}
\vspace{-0.55cm}
\caption{Visual comparison of RIFE (left) and HSTR-Net (right) on Visdrone images.}
\label{fig:comparison}
\end{figure}
\vspace{-0.2cm}
Figure \ref{fig:comparison} illustrates reconstructed frames inferred from highly rotational drone video feeds. As seen from the figure, RIFE\cite{huang2020rife} cannot compensate for the rotation of camera and creates distorted and blurry output. HSTR-Net successfully incorporates LR frames features and reconstructs visually much better result.

\vspace{-0.3cm}
\section{Conclusion}
\label{head: conclusion}
\vspace{-0.2cm}
In this paper, we propose a reference-based end-to-end trainable network named HSTR-Net for temporal frame interpolation. We demonstrate that LR frames are useful for HR frame reconstruction, especially when there is large and complex motion in the scene.  We achieve state-of-the-art results by replacing non-trainable warping with trainable deformable convolution layer and show that using trainable modules generates more accurate results. The models in this work are trained and tested in synthetically generated datasets.
%We have shown that better results can be obtained utilizing dual video feeds even though HSTRNet is not %developed for dual cameras. 
As future work, we plan to employ the proposed network on an actual real-life dual camera setup.
\vspace{-0.2cm}
\bibliographystyle{IEEEbib}
\bibliography{refs}
\end{document}